# Pixel precise unsupervised detection of viral particle proliferation in cellular imaging data

Birgitta Dresp-Langley[1], ICube Lab UMR 7357 CNRS-Strasbourg University, FRANCE
**John Mwangi Wandeto**[2], Department of Information Technology, Dedan Kimathi University, **KENYA**

[2]**corresponding author**: ndetos@gmail.com
P.O. Box 12305 – 10109, Nyeri Kenya

**Abstract**

Cellular and molecular imaging techniques and models have been developed to characterize single stages of viral proliferation after focal infection of cells *in vitro*. The fast and automatic classification of cell imaging data may prove helpful prior to any further comparison of representative experimental data to mathematical models of viral propagation in host cells. Here, we use computer generated images drawn from a reproduction of an imaging model from a previously published study of experimentally obtained cell imaging data representing progressive viral particle proliferation in host cell monolayers. Inspired by experimental time-based imaging data, here in this study viral particle increase in time is simulated by a one-by-one increase, across images, in black or gray single pixels representing dead or partially infected cells, and hypothetical remission by a one-by-one increase in white pixels coding for living cells in the original image model. The image simulations are submitted to unsupervised learning by a Self-Organizing Map (SOM) and the Quantization Error in the SOM output (SOM-QE) is used for automatic classification of the image simulations as a function of the represented extent of viral particle proliferation or cell recovery. Unsupervised classification by SOM-QE of 160 model images, each with more than three million pixels, is shown to provide a statistically reliable, pixel precise, and fast classification model that outperforms human computer-assisted image classification by RGB image mean computation. The automatic classification procedure proposed here provides a powerful approach to understand finely tuned mechanisms in the infection and proliferation of virus in cell lines *in vitro* or other cells.

**Introduction**

Viruses are cellular parasites that do not have a metabolism of their own. To proliferate, they must infect healthy host cells. To achieve this, virus particles are able to recognize and to bind specific receptor molecules on the cytoplasmic membranes of host cells, a process that is known as attachment. This interaction between viral particles and a host cell's membrane surface is mediated by proteins embedded within the viral envelope [1]. For example, the binding of viral envelope proteins to a host cell's surface structure in the case of the human immunodeficiency virus (HIV), which is a coronavirus, involves interaction between the surface protein gp120 and the CD4 receptor [2], a polypeptide that occurs almost exclusively in the membranes of T-helper cells (lymphocytes) and macrophages. This first step of the HIV replication cycle critically determines the ability of HIV to degrade the human immune system. In other cases, viral proteins may bind to cellular structures found on a larger variety of cell types.

Various tools for cellular and molecular imaging have been developed to characterize single intracellular stages of viral proliferation [1,3]. The high resolving power of electron microscopy (EM) permits studies at nanometer scale, providing direct images of viral proliferation (virion budding) on host cell membrane surfaces for diagnosis and research [4]. Scanning electron microscopy (SEM) in particular facilitates single cell analysis and comparison between cells by visualizing ultrastructual changes on membrane surfaces and structures harboring viral replication sites correlated with the progression or remission of the infectious process [2,4,5].

Immunocytochemistry [5,6] and cell viability imaging by color staining and other labeling techniques allow the study of spatial and temporal dynamics of virus spreading on host cell surfaces *in vitro*. Haseltine et al. [7] developed an imaging method that permits tracking the spread of focal infections using a combination of immunocytochemical labeling and step-by-step digital imaging. In their study, baby hamster kidney (BHK) cells were seeded on six well plates, grown as confluent monolayers and covered with a thin layer of agar. After piercing a small orifice in the agar, cell layers were infected by injecting 5 µl of virus VSVN1 inoculum at $1.6 \times 10^7$ infectious particles (i.e. a multiplicity of infection of 20) as described in [6,7]. Cells were subsequently fixed, and immunofluorescence labeled with an antibody against a viral glycoprotein on and within the infected cells. Images of (Figure 1) were then acquired by epifluoresence microscopy at low magnification using a high-resolution monochrome digital camera.

The temporal dynamics of infection spreading guided the iterative modeling steps that led to a mathematical reconstruction of its relevant characteristics. Haseltine et al.'s [7] model approach thus was aimed at facilitating the interpretation of experimental labeling data by highlighting critical spatial and temporal aspects of viral infection and proliferation on a host cell. Any technique for the determination of *in vitro* cell changes and cell viability combined with any imaging technique or advanced imaging model involves human visual classification and/or interpretation of the image material, which may involve guesswork when the spatiotemporal uncertainty of the image contents is high. Under such conditions it is difficult to rule out any subjectivity of the analyst [5]. To ensure high-quality decision making in the evaluation of larger amounts of required. In addition, affordable precision software for an automatic classification of cell imaging data should combine high accuracy of classification with further advantages relative to speed, objectivity, and reproducibility of the quantifications. In our previous studies, we exploited functional properties such as sensitivity to spatial extent, intensity, and color of local image contrasts of the quantization error in the output of a Self-Organizing Map (SOM-QE) for unsupervised image classification as a function of the finest, often visible, clinically or functionally relevant variations in local contrast contents [8-15]. Regarding cell imaging data, the SOM-QE was successfully employed for fast unsupervised classification of SEM images of CD4 T-cells with varying extent of ultra-structural surface signals correlated with surface-localized single HIV-1 viral particle infection. It was shown that SOM-QE permits a fast automatic classification of sets of such SEM images as a function of ultra-structural signal changes that are invisible to the human eye. As pointed out in our previous studies, any imaging method currently available has its own limitations, including for SEM. It is important that all images from a series with presumed functional variations submitted to pixel precise analysis are identically scaled. SEM is meant to provide a 3D model of ultrastructural cell properties, however, the resulting images that are subject to any further analysis are in 2D, with a z-axis projecting into virtual depth. This leaves additional room for viewpoint-related errors, and it is necessary to ensure that images from a given time series are not only identically scaled in 2D, but are taken from one and the same 3D view of the cell, or any other structure, when submitted to analysis.

SOM-QE is also a powerful tool for the automatic classification of double-color-staining based cell viability data, as shown previously in 96 image simulations [9]. SOM-QE consistently detects the smallest spatial changes in any of the local color signals [14,1]. These may reflect a systematic increase or decrease in theoretical cell viability below the expert visibility threshold. Signal detection experiments [9, 11] have shown that clinically significant changes in the color or contrast of less than 10 pixels between two images of a series with clinical significance are impossible to detect, even by temporally unlimited expert visual inspection. SOM-QE detects any such changes in a consistent and statistically reliable manner, as shown in proof-of-concept studies demonstrating the selectivity of the metric to the spatial extent, the contrast intensity, the contrast polarity and the color of single-pixel changes in random-dot images containing millions of image pixels [12-15].

In the present study, we use the previously published cell imaging model by Haseltine et al. [7], described in all necessary detail here above. It is shown that the fast automatic classification by SOM-QE gives access to the earliest potential stages of viral propagation in host cells, which are detectable neither by visual inspection, nor by the imaging model *per se*. The imaging model described in [7] was aimed at providing an age-segregating model for the time course, in terms of hours or days, of viral particle proliferation following focal infection of the cell monolayer. The original images from the reference study [7] quantize the sum of infected and dead cell concentrations, a continuous variable, onto an integer-valued intensity scale. A high resolution computer software reconstruction of one of the images from the reference study (see Figure 1), taken at a particular moment in time in the course of *in vitro* monolayer infection, was used as the ground truth image here. The model copy derived from the initial image is then used to simulate image sequences in theoretical time corresponding to seconds or minutes of the smallest possible infectious progression/recession of the virus in the cell monolayer. Viral particle growth with time is simulated by a one-by-one increase, across images. Black pixels indicate dead, gray pixels represent slightly infected, white pixels represent living cells in the original and the hypothetical image model.

The image simulations are then submitted to unsupervised learning by a Self-Organizing Map (SOM) and, as described previously [8-15], the Quantization Error in the SOM output (SOM-QE) is used for the automatic classification of the image simulations as a function of the represented extent of viral particle proliferation or cell recovery.

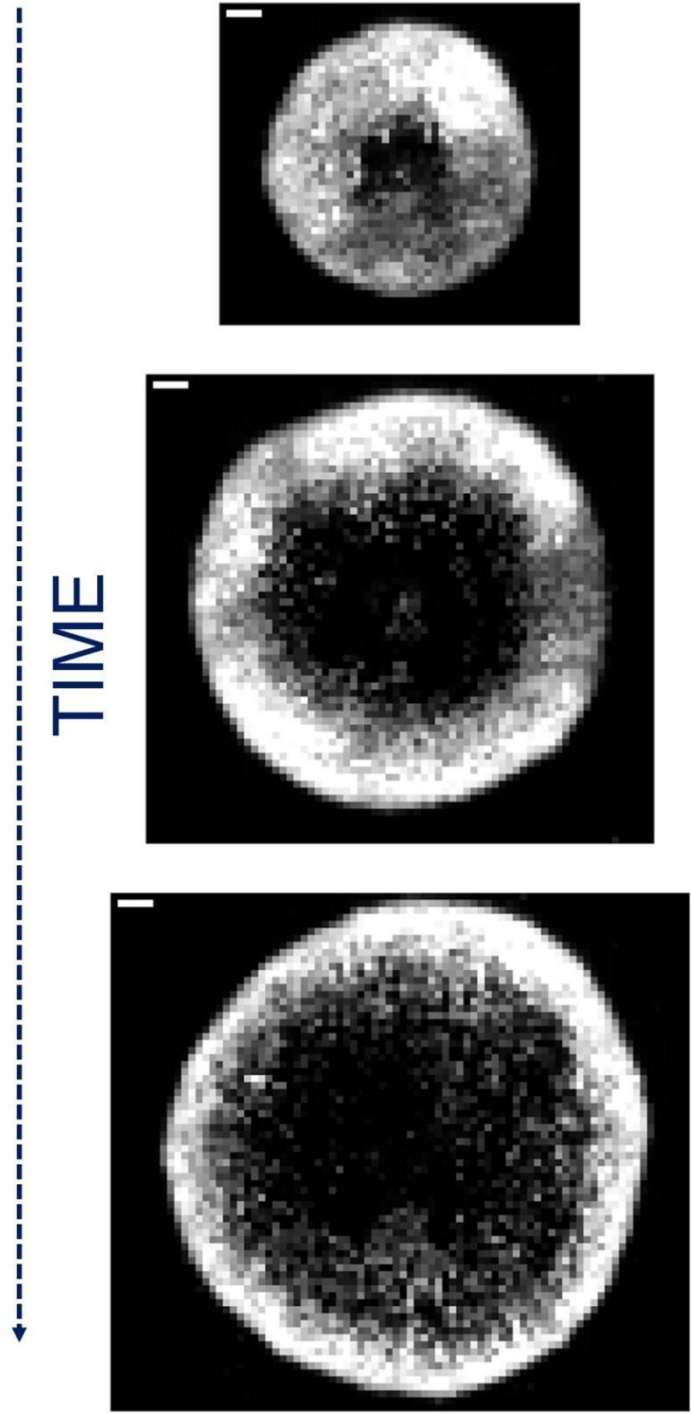

**Figure 1:** Example of a time-based imaging model (reproduced, modified, and shown here for illustration only) from the study by Haseltine et al. [7], where images of infected cells were generated at different times after focal infection of a cell monolayer. As described in [7], after focal infection *in vitro,* antibody labeling for viral glycoprotein, and detection by immunofluorescence microscopy, digital camera images of the infected monolayer region (circular image region) were generated. Contrast levels in the circular region represent relative amounts of living cell matter (white image pixels), progressing infection (image pixels in varying shades of gray), and dead cell matter (black image pixels). Images from top to bottom

here correspond to times between 48 (top), 72 (middle) and 90 hours (bottom) of infection. Scale bars represent 1mm. The model image from which all image simulations for this study here were drawn is the top image of this figure.

**Materials and Methods**

*Image simulations*

A total number of 160 images were computer generated, using the single-pixel-level control functions in ADOBE Photoshop7, on the basis of a virtual copy of the first of the original images shown in Figure 1, which is a low resolution version of the high resolution ground truth image, shown in Figure 2, that was reconstructed for this study here prior to any further image simulations. Figure 2 shows the image state (ground truth) before any changes simulating further viral particle growth/recession were implemented. 159 further images were drawn directly from this high resolution computer model image to simulate the finest possible (i.e. observable in the image) viral particle growth or cell recovery in time. It took 30 minutes to reconstruct a model copy of the reference image, and about one hour to generate the 160 image simulations of single pixel change. In an experimental cell study, changes reflected by the small single pixel changes in our high resolution image simulations could occur within minutes after *in vitro* infection.

Viral particle growth was simulated by replacing, one by one, image pixels indicating living cell matter (white) by pixels indicating different levels of infection (shades of gray) or dead cell matter (black). In a hypothetical situation in which a specific cell treatment could enable cells to resist cell death and recover from infection, cell recovery in time was simulated by a one-by-one increase, across images, in light pixels representing living cell matter. Image pixels indicating dead cell matter (black) were replaced by pixels indicating progressive levels of infection (shades of gray) or living cell matter (white). All 160 images had identical size (1906x1794) with a total number of 3 419 364 pixels. RED-GREEN-BLUE (RGB) values for the different gray levels of image pixels were [0, 0, 0], referred to as 'black' here, [13, 13, 13], [38, 38, 38], [64, 64, 64], referred to as 'dark gray' here, [89, 89, 89], [127, 127, 127], referred to as 'medium gray' here, [191, 191, 191], [217, 217, 217], referred to as 'light gray' here, [242, 242, 242], and [255, 255, 255], referred to as 'white' here.

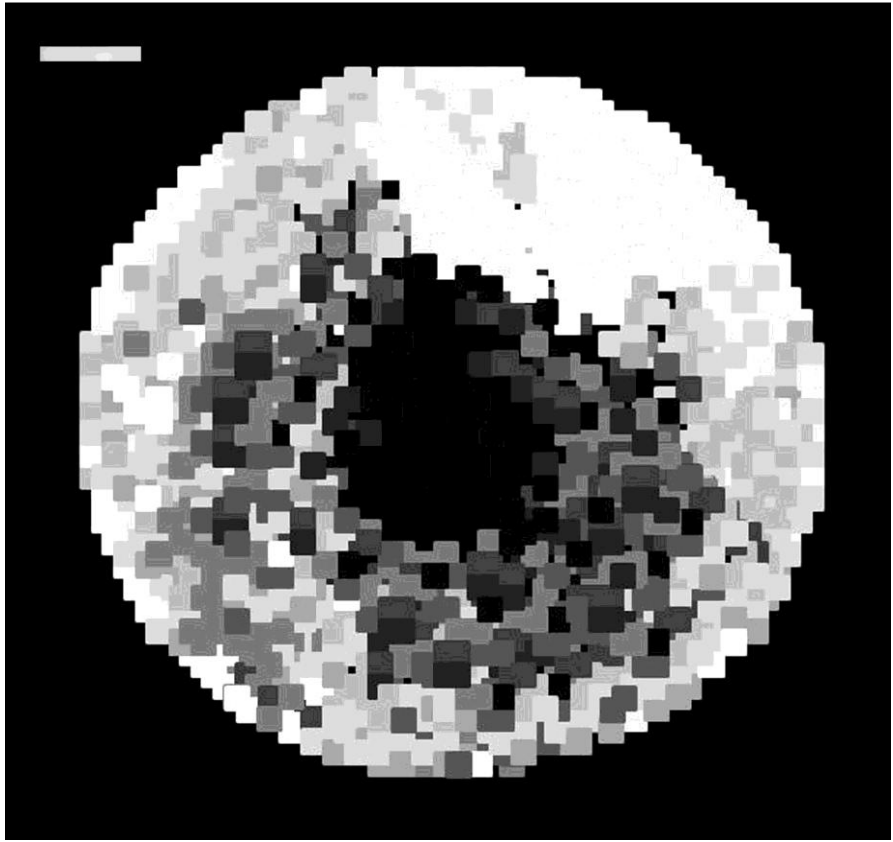

**Figure 2:** Computer generated image model reconstruction based on an original image from experimentally obtained cell imaging data [7]. The model possesses the same clinically relevant image data variations as the original (Figure 1) at the same scale. A total of 160 images were computer generated, in *ADOBE Photoshop7,* from this model reconstruction. Single-pixel RGB increments simulate theoretical image data for hypothetical viral particle growth/recession in time. This is achieved by replacing, one by one, pixels indicating living cell matter (white) by pixels indicating different levels of infection (shades of gray) or dead cell matter (black). Cell recovery after a hypothetical treatment was simulated by replacing, one by one, image pixels indicating dead cell matter (black) by pixels indicating different levels of infection (shades of gray) or living cell matter (white).

*Self-Organizing Map (SOM) and Quantization Error (QE)*

The Self-Organizing Map (a prototype is graphically represented in Figure 3, for illustration) may be described formally as a nonlinear, ordered, smooth mapping of high-dimensional input data onto the elements of a regular, low-dimensional array [16]. It is assumed that the set of input variables is definable as a real vector $x$, of n-dimension. A parametric real vector $m_i$ of n-dimension is associated with each element in the SOM. Vector $m_i$ is a model and the SOM is therefore an array of models. Assuming a general distance measure between $x$ and $m_i$ denoted by *d(x,m$_i$)*, the map of an input vector *x* on the SOM array is defined as the array element *m$_c$* that matches best (smallest *d(x,m$_i$)*) with *x*. During the learning process, the input vector *x* is compared with all the *m$_i$* in order to identify *m$_c$*. The Euclidean distances ||*x-m$_i$*|| define *m$_c$*. Models topographically close in the map up to a certain geometric distance, indicated by *h$_{ci}$*, will activate each other to learn something from their common input *x*. This results in a local relaxation or smoothing effect on the models in this neighborhood, which in continuous learning leads to global ordering. SOM learning is represented by the equation

$$m(t+1) = m_i(t) + \alpha(t)h_{ci}(t)[x(t) - m_i(t)] \qquad (1)$$

where $t = 1,2,3...$ is an integer, the discrete-time coordinate, *h$_{ci}$(t)* is the neighborhood function, a smoothing kernel defined over the map points which converges towards zero with time, $\alpha(t)$ is the learning rate, which also converges towards zero with time and affects the amount of learning in each model. At the end of the *winner-take-all* learning process in the SOM, each image input vector *x* becomes associated to its best matching model on the map *m$_c$*. The difference between *x* and *m$_c$*, ||*x-m$_c$*||, is a measure of how close the final SOM value is to the original input value and is reflected by the quantization error, QE. The average QE of of all *x* (X) in an image is given by:

$$QE = 1/N \sum_{i=1}^{N} \lVert X_i - m_{c_i} \rVert \qquad (2)$$

where N is the number of input vectors *x* in the image. The final weights of the SOM are defined by a three dimensional output vector space representing each R, G, and B channel. The magnitude as well as the direction of change in any of these from one image to another is reliably reflected by changes in the QE.

The code used for implementing the SOM-QE is available online at:

*https://www.researchgate.net/publication/330500541_Self-organizing_map-based_quantization_error_from_images*

The SOM training process consisted of 1 000 iterations. The SOM was a two-dimensional rectangular map of 4 by 4 nodes, hence capable of creating 16 models of observation from the data. The spatial locations, or coordinates, of each of the 16 models or domains, placed at different locations on the map, exhibit characteristics that make each one different from all the others. When a new input signal is presented to the map, the models compete and the winner will be the model the features of which most closely resemble those of the input signal. The input signal will thus be classified or grouped in one of models. Each model or domain acts like a separate decoder for the same input, i.e. independently interprets the information carried by a new input. The input is represented as a mathematical vector of the same format as that of the models in the map. Therefore, it is the presence or absence of an active response at a specific map location and not so much the exact input-output signal transformation or magnitude of the response that provides the interpretation of the input. To obtain the initial values for the map size, a trial-and-error process was implemented. Map sizes larger than 4 by 4 produced observations where some models ended up empty, which meant that these models did not attract any input by the end of the training. As a consequence, 16 models were sufficient to represent all the fine structures in the image data. Neighborhood distance and learning rate were constant at 1.2 and 0.2 respectively. These values were obtained through the trial-and-error method after testing the quality of the first guess, which is directly determined by the value of the resulting quantization error ; the lower this value, the better the first guess. It is worthwhile pointing out that the models were initialized by randomly picking vectors from the training image, called here the "original image". This allows the SOM to work on the original data without any prior assumptions about a level of organization within the data. This, however, requires to start with a wider neighborhood function and a bigger learning-rate factor than in procedures where initial values for model vectors are pre-selected [17]. The approach is economical in terms of computation times, which constitutes one of its major advantages for rapid change/no change detection on the basis of even larger sets of image data prior to any further human intervention or decision making.

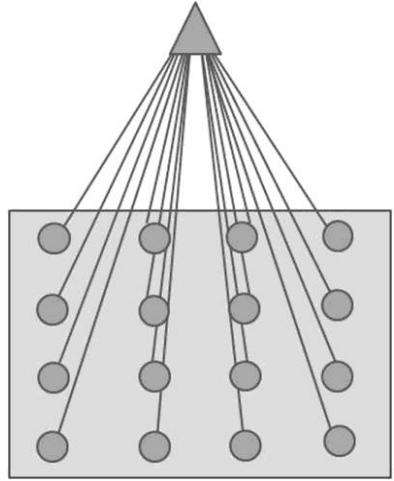

**Figure 3**: Representation of the SOM prototype with 16 models, indicated by the filled circles in the grey box. Each of these models is compared to the SOM input in an unsupervised winner-take-all learning process. Here in this study, the input vector corresponds to the RGB image pixel space. The model in the map best matching the SOM input will be a winner, and the parameters of the winning model will change towards further approaching the input. Parameters of models within close neighborhood of the winning model will also change, but to a lesser extent than those of the winner. At the end of training, each input space is associated with a model in the map. The difference between input vector and final winning model determines the quantization error (QE) in the SOM output.

*Experimental procedure*

The 160 images simulating viral particle growth by progressive one-by-one replacement, in the region inside the circle of infection, of white ("living cell matter") pixels by light gray, medium gray, dark gray or black ones, or cell recovery, following a hypothetical treatment specifically designed to enable cells to recover from the effects of infection, by progressive one-by-one replacement of black ("dead cell matter") pixels by light gray, medium gray, dark gray or white ones were fed into a single SOM. The training image for the SOM prior to analysis was the model image 'before any pixel change', shown in Figure 2. After unsupervised SOM learning of the training image followed by SOM analysis of all images, the SOM-QE output was written into a data file. Further steps generate output plots of SOM-QE, where each output value is associated with the corresponding input image. The output data are then plotted in increasing/decreasing orders of SOM-QE magnitude as a function of the corresponding image variations (automatic image classification). The computation time of SOM analysis of each of the 160 images was about 12 seconds per image. The precision of the SOM-QE image classification was then compared with that of human computer-assisted classification of the same image data on the basis of the RGB Mean in *Image-J*, an open access image analysis tool available online at: *https://imagej.nih.gov/ij/* .

**Results**

The SOM-QE output data of automatic image classification by unsupervised SOM learning and the output data of human computer-assisted image classification in terms of the RGB image means were plotted as a function of the magnitude of image pixel changes simulating progressive, i.e. increasing spatial extent of, viral proliferation or progressive, i.e. increasing spatial extent of, cells recovering from higher levels of infection, simulated here, in consistency with the experimental cell imaging approach described in [7], by different shades of gray. These results are shown here in Figure 4 for images simulating progressive viral proliferation, and in Figure 5 for images simulating progressive cell recovery from infection. The X axes of the graphs indicate the order of the images under the assumption that pixel changes in the image simulations in that order reflect the underlying viral proliferation/cell recovery process in time. The Y axes give the units of each metric (RGB Mean, SOM-QE) used to detect the changes in the images.

*Descriptive analysis*

The data in Figure 4 (bottom graph) show that the SOM-QE consistently and reliably classifies the cellular changes in the images by signaling a linear decrease in the QE as a function of increasing the number of gray or black image pixel replacing white image pixels, simulating progressive viral proliferation in the infected cell monolayer region. The RGB image means (Figure 4, top and middle graphs) from human computer-assisted image classification *(ImageJ)* does not allow to classify the image data in a statistically reliable way. In other words, the image mean fails to detect the finer pixel changes in the images. While the SOM-QE is highly and consistently sensitive to all changes simulating different intensity levels of infection from light gray to black pixel tones, with a marked capacity to discriminate between changes in light gray and black (Figure 4, bottom graph), the RGB mean fails completely to detect any changes in the number of light gray pixels replacing white ones (Figure 4, middle graph on right) simulating local infection progression in the images.

The data in Figure 5 (bottom graph) show that the SOM-QE consistently and reliably classifies the cellular changes in the images by signaling a linear increase in the QE as a function of increasing number of gray or white image pixel replacing black image pixels, simulating progressive cell recovery after a hypothetical treatment, i.e. increasing amount of healthier living cell tissue. The RGB image means (Figure 5, top and middle graphs) from human computer-assisted image classification *(ImageJ),* again, do not permit to reliably classify the image data. Again, the image mean does not detect the finer pixel changes in the images, while the SOM-QE detects them reliably.

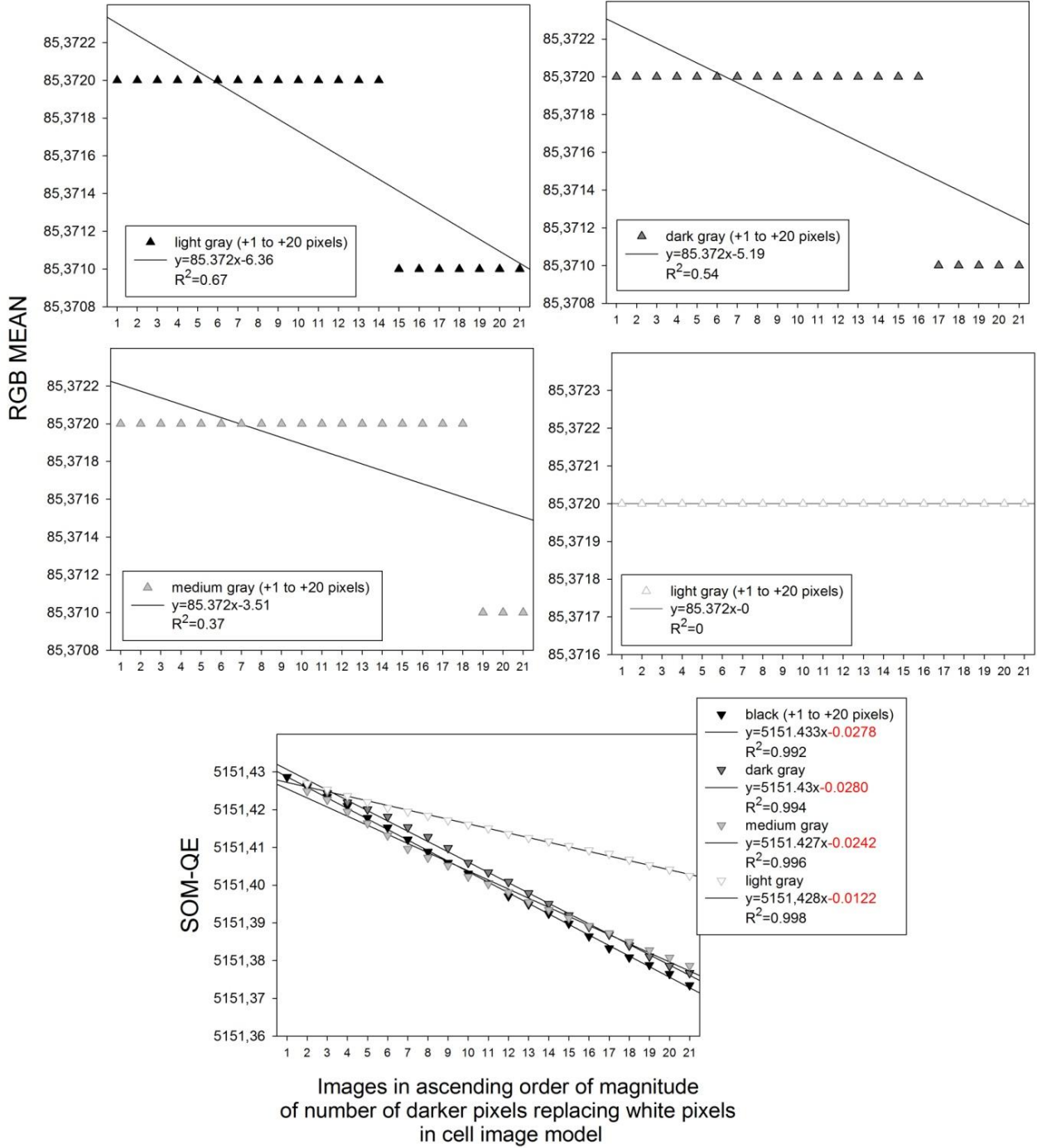

Images in ascending order of magnitude
of number of darker pixels replacing white pixels
in cell image model

**Figure 4**: Graphic representations of the SOM-QE classification data (bottom graph) by comparison with the data from the human computer-assisted (*ImageJ*) classification in terms of RGB image means (top and middle graphs). The output data are plotted here as a function of the magnitude of image pixel changes simulating progressive viral proliferation for varying levels of infection. These are simulated in the images by darker pixels indicating infection progression towards cell death replacing white image pixels indicating living cell tissue.

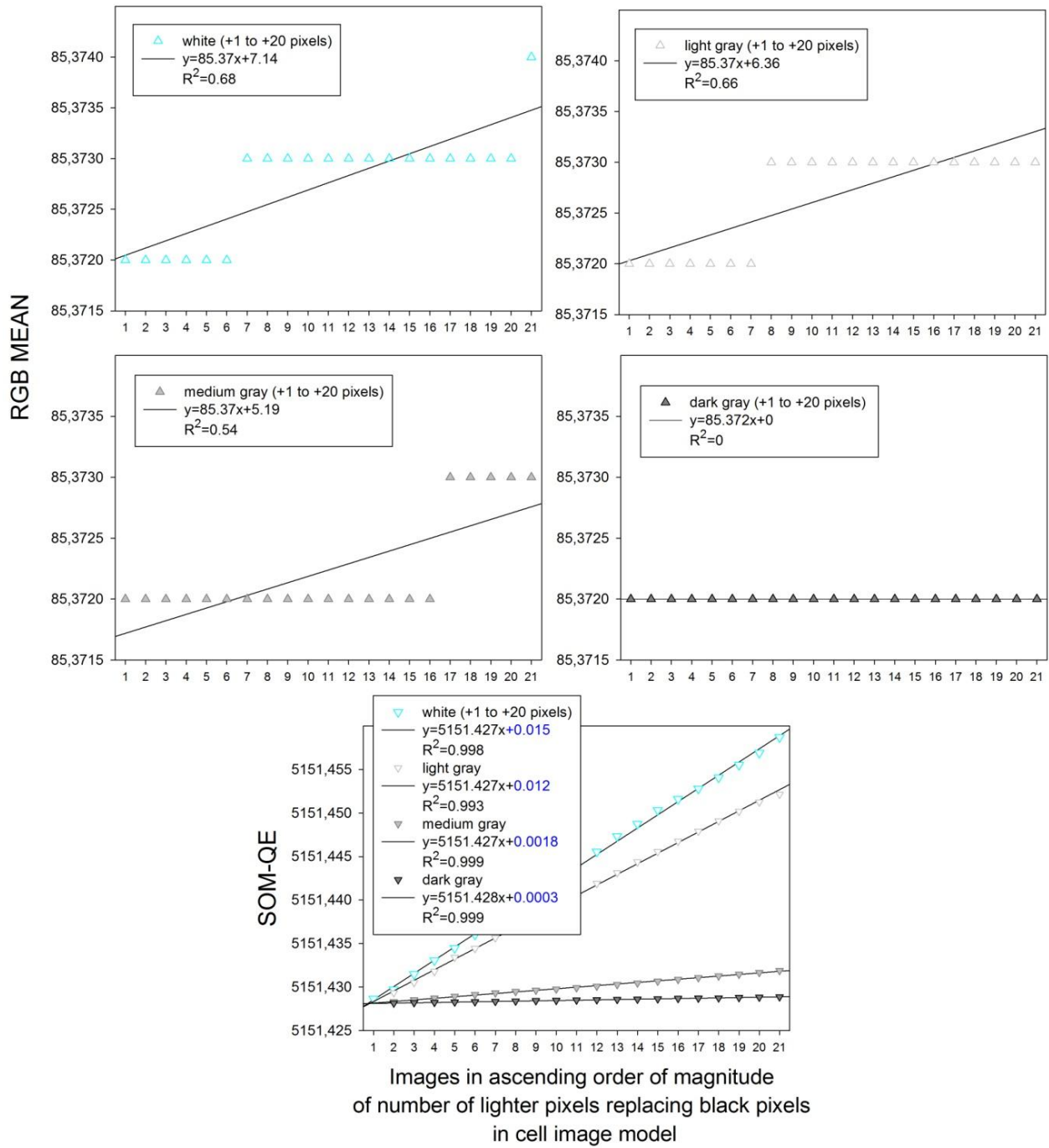

**Figure 5**: Graphic representations of the SOM-QE classification data (bottom graph) by comparison with the data from the human computer-assisted (*ImageJ*) classification in terms

of RGB image means (top and middle graphs). The output data are plotted here as a function of the magnitude of image pixel changes simulating progressive (i.e. increasing spatial extent of) cell recovery for varying levels of infection. These are simulated in the images by lighter image pixels replacing black image pixels indicating dead cell tissue.

again, highly and consistently sensitive to all changes simulating different intensity levels of cell tissue recovery, from light gray to white pixel tones, with a marked capacity to discriminate between all gray pixel and white pixels changes (Figure 5, bottom graph), the RGB mean fails completely to detect any changes in the number of dark gray pixels replacing black ones (Figure 5, middle graph on right), simulating low-level cellular recovery here, in the images.

*Statistical analysis*

The SOM-QE data and the RGB image means plotted in Figures 4 and 5 were submitted to linear regression analysis to further quantify the reliability of unsupervised SOM-QE classification by comparison with the RGB mean. The linear equations of the fits, with numerical values for slopes (a) and intercepts (b) of the functions, are displayed in the Figures 4 and 5. The goodness of these fits are assessed and compared on the basis of two statistical criteria: 1) the normality test (Shapiro-Wilk herein), which yields a probability criterion for the claim that the output distribution, the SOM-QE and/or the RGB mean distributions herein, of a given classification model consistently reproduces the theoretically assumed normal distribution and 2) the regression coefficient $R^2$, where $R^2 = 1$ when the output data distribution is a perfectly linear model of the classification input data.

**Table1.** Linear regression analysis of the SOM-QE and RGB Mean classification data

| Normality test (Shapiro-Wilk) | | Linear regression coefficient ($R^2$) | | *simulated pixel precise cell infection progression or cell recovery* |
|---|---|---|---|---|
| **SOM-QE** | **RGB Mean** | **SOM-QE** | **RGB Mean** | *hypothetical infection levels* |
| .42 *passed* | .05 *failed* | .99 | .67 | *black replacing white* |
| .48 *passed* | .05 *failed* | .99 | .54 | *dark gray replacing white* |
| .53 *passed* | .05 *failed* | .99 | .32 | *medium gray replacing white* |
| .84 *passed* | .05 *failed* | .99 | 0 | *light gray replacing white* |
| | | | | *hypothetical recovery levels* |
| .42 *passed* | .05 *failed* | .99 | .68 | *white replacing black* |
| .47 *passed* | .05 *failed* | .99 | .66 | *light gray replacing black* |
| .60 *passed* | .05 *failed* | .99 | .54 | *medium gray replacing black* |
| .74 *passed* | .05 *failed* | .99 | 0 | *dark gray replacing black* |

The results of the normality tests and the regression analyses ($R^2$), shown here in Table 1, show that the SOM-QE distributions or image classification data satisfy the normality criterion in all test cases. The regression coefficients $R^2$ of the linear fits are .99 in all cases, indicating that the SOM-QE classification model of the image data is statistically reliable and provides a quasi-perfect detection model for the changes in the image input simulating pixel precise viral proliferation or recovery. Conversely, the RGB-Mean distributions or image classification data fail the normality criterion in all test cases, and the regression coefficients $R^2$ of the linear fits are <.70 in all cases, indicating that the RGB-Mean fails to reliably classify the changes in the image input simulating pixel precise viral proliferation or recovery. For a direct statistical comparison, we used a t-test to assess the statistical significance of the difference between the regression coefficient distributions relative to the linear fits for the SOM-QE and the RGB Mean. The result of this test signals a statistically significant difference between the distributions (t (1,14) = 5.544; p<.001).

**Discussion**

The results from this study show that the unsupervised, self-organizing map-based automatic classification by SOM-QE [8-15] of cellular imaging models simulating viral proliferation in the cell [7], or cellular recovery after a specific hypothetical *in vitro* cell treatment in time with a single-pixel precision provides a statistically highly reliable classification model that by far outperforms human computer-assisted image classification in terms of the RGB image mean, for example, as demonstrated here. The linear models for SOM-QE increase or decrease as a function of local changes in single image pixel contrast reproduce the previously shown fine sensitivity of this neural network metric to changes in the contrast polarity and/or intensity of single pixels in several millions of image pixels [8-15]. The theoretical model images that were fed into SOM-QE analysis here each consist of more than three million pixels. The hypothetical timescale of the cell changes simulated by single-pixel changes between these high resolution images, for either viral progression or cell recovery, in either a single cell or in a monolayer of cells, may be estimated in seconds/minutes. With increasingly high resolution camera solutions, and images from a constant camera position taken every ten minutes and fed directly into a computer for SOM-QE analysis, cell biologists could become able to detect significant changes within cell monolayers or single cells within unprecedentedly short delays.

To provide a time course model for viral proliferation in terms of hours/days, the authors of the reference study [7] further reduced the complexity of information in their experimentally derived images, which had a considerably poorer resolution than our model image simulations here, by partitioning them into blocks of 20x20 pixels, which is not very precise, then averaging the pixels contained in each block. This reduction in image complexity was necessary to reduce the total number of pixels while retaining prominent features of the infection spread for further analysis by their own mathematical model. In the case of their largest image, this represented a reduction from roughly two million to five thousand pixels in their study [7]. Image reduction to deal with too much complexity always runs a risk of loss of potentially relevant information. Since SOM-QE provides pixel precision analysis of very large images, image data reduction is not required to reliably classify even the smallest local changes.

SOM-QE can also be applied to selected, particularly relevant image sections, or regions of interests (ROI), for analysis of single pixel changes whenever their spatial location in an image is a critical factor, as shown in some of our previous work [12]. Combining immunohistochemical data with digital imaging is definitely the way to go in the future. The exploitation of digital images in cancer diagnosis [18,19,20], for example, optimized by image segmentation methods capable of detection ROI automatically prior to further classification by SOM-QE, can provide objective analyses that make the difficult task of diagnosis more reliable and less time consuming for the human expert. Finally, in the current context of pandemic explosion of SARS-CoV, a particular class of coronavirus, there is hope that cellular imaging models, like the one described in [7], which inspired this study here, that account for *in vitro* coronavirus entry and proliferation mechanisms in cell lines or other cultured cells will allow for a better understanding of the infection process [21]. Rapid, non-expensive, and swiftly implemented automatic image classification methods like the one proposed here could support the fast development of novel treatment strategies.

**Conclusions**

Digital image pixel RGB based models of viral proliferation after focal infection *in vitro*, or cell recovery in response to a specific treatment, may contain millions of image pixels that cannot be processed visually even by an experienced expert. The unsupervised self-organizing map based SOM-QE approach [8-15] is shown to provide a fast and statistically highly reliable classification model for the finest single pixel changes in image models containing several millions of pixels, simulating viral proliferation or cell recovery in terms of viral level reduction by progressive image pixel changes in contrast polarity and/or intensity. The classification method outperforms human computer-assisted image classification, is fast and economic, and can be applied to large imaging data prior to further mathematical modeling.